# Precise classification of low quality G-banded Chromosome Images by reliability metrics and data pruning classifier

**Abstract**  Mojtaba Moattari   Independent Researcher   Moatary.@gmail.com   Jan 2022

In the last decade, due to high resolution cameras and accurate meta-phase analyzes, the accuracy of chromosome classification has improved substantially. However, current Karyotyping systems demand large number of high quality train data to have an adequately plausible Precision per each chromosome. Such provision of high quality train data with accurate devices are not yet accomplished in some out-reached pathological laboratories. To prevent false positive detections in low-cost systems and low-quality images settings, this paper improves the classification Precision of chromosomes using proposed reliability thresholding metrics and deliberately engineered features. The proposed method has been evaluated using a variation of deep Alex-Net neural network, SVM, K-Nearest-Neighbors, and their cascade pipelines to an automated filtering of semi-straight chromosome. The classification results have highly improved over 90% for the chromosomes with more common defects and translocations. Furthermore, a comparative analysis over the proposed thresholding metrics has been conducted and the best metric is bolded with its salient characteristics. The high Precision results provided for a very low-quality G-banding database verifies suitability of the proposed metrics and pruning method for Karyotyping facilities in poor countries and low-budget pathological laboratories.

## 1   Introduction

One of the ways to study and diagnose birth-defects and biological disorders is through using Cytogenetics. This branch of science endeavors to analyze chromosome shapes and patterns to find out common defects. The methods used for such analyzes includes G-Banding, Fluorescent In-Situ Hybridization (FISH), Comparative Genomic Hybridization (CGH) and Chromosome-specific unique-sequence probes [27]. While Molecular Cytogenetics methods are effective in biological disorders, they do not necessarily manifest specific chromosome defects. FISH methods, though having higher accuracy results in stains, are costly and unable to identify all chromosome abnormalities. Being temporary in sustaining fluorescence detector, they demand higher provision effort and substance supply that might not be affordable for some countries. Furthermore, detecting some abnormalities implies having G-banding technique involved and not merely using stains.

Fig 1 shows a meta-phase for a male human extracted by G-banding technique. In part a, the pool of unsorted chromosomes are shown, while part b manifests them in a sorted way for diagnosis purposes and defect detections. In this case, the karyotyped person is healthy.

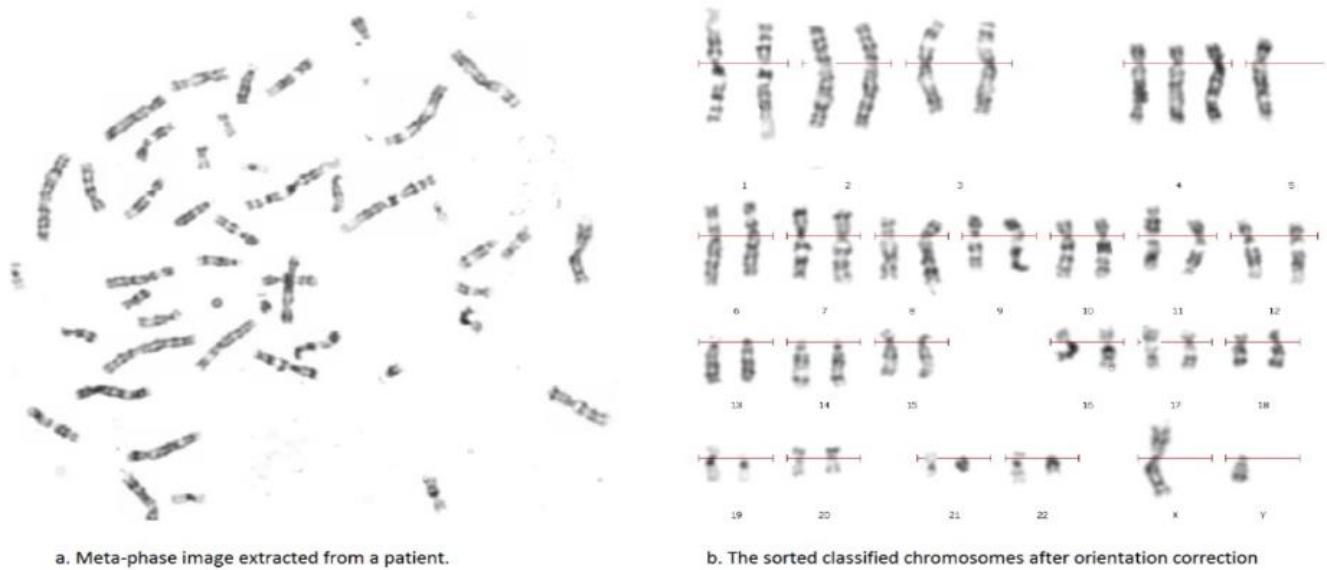

**Fig 1**   A sample from Meta-phase database and its reordering after correction by the operator.

The systems generally used for sorting chromosomes, are either of two types, automated [8, 28], and semi-automated [4]. Automated Karyotyping methods demand high number of train data (>3000 per label), to generalize well to each chromosome model. Their high sensitivity to noise and input changes, causes them to demand high quality imaging tools with highly accurate meta-phase analysis, making such approach impractical for low-budget pathological laboratories especially in poor countries. On the other hand, semi-automated softwares facilitate cooperation between expert and software to segment, detect and sort parts of the chromosomes leaving the remnant for the operator. Not only is this approach more plausible in low-quality settings, as the program provides the operator to modify and correct results, but also it avoids the possible mistakes in automated software that impose extra job for rechecking all the chromosomes. It makes development of highly reliable semi-automated classifiers of substantial importance in Metaphase Analysis. To improve reliability and certainty, the software has to discern where the detection is not precise enough concerning the context of input pattern it fed on. In this paper, new set of metrics have been proposed to tackle the reliability of classification and give classifiers the choice to return no labels during uncertainty. To the best of authors' knowledge, no methods and metrics so far have been proposed for improving semi-automated classification reliability and Precision in chromosome classification. Therefore, to evaluate the proposed approach, the Precision with and without the implemented reliability metrics has been compared.

Varieties of methods have been used for classifying chromosomes, but in different Cytogenetics methods. Among the most commonly-used classifiers, statistical models and Artificial Neural Networks (ANNs) are proved more effective, especially compared to Fuzzy Inference Systems (FIS) and linear classifiers [20, 13]. In 1989, Britto used multiple segments of each chromosome image to be passed through its respective classifier, and performed decision by voting over aggregated results [3]. Cosio trained a group of 10-feature instances in 2001 which contained a variety of features including morphological, photometric and image context information inside [6]. The network, being of the first comprehensive usages of ANN in chromosome classification contexts, trained the network in a supervised manner and the final model could outperform all the previous methods. The generality of metrics facilitates their implementation on new classifiers and settings.

Delshadpour trained an ANN perceptron over varieties of morphological, and tuned number of neurons to improve accuracy from 77.8% to 85.3% [7]. Other efforts also for classification were conducted by Maximum Likelihood (ML) approach after transforming images to a polar coordinate space and Median Filtering them as a denoising method [3]. Masecchia classified based on Dictionary Learning concepts in which chromosomes would be analyzed to form class-specific components that code each instance to its corresponding sparse representation [18]. Various Deep architectures are proposed and improved with the cost of long training time and necessity of high quality images [14, 26]. Pardo in 2018, combined FISH segmentation and classification in a unified framework called Semantic Segmentation. However, evaluated dataset was FISH database rather than G-Banded images as provided in this paper [23]. The implemented model was a Convolutional Neural Network (CNN).

However, none of these methods tried to improve precision using thresholding metrics. Generalized score-thresholding metrics open new opportunities for reliability improvement of previous models and guaranty to improve precision on any model case including deep structures. Furthermore, so far methods have focused exclusively on improving certainty and reliability of high-resolution semi-automated chromosome classification. However, this paper's focus is on finding precise models over low-quality images, i.e. exceeding 90% generally and getting over 97% in some cases. To the best of authors' knowledge, for image resolutions behind 70x70 pixel, no methods provided Precisions over 90%. Overall, evaluation results of chromosome classification mentioned in Section 3 suggest that designation of accurate classifiers over low-resolution data can lead to precise semi-automated softwares and their beneficial roles in the laboratories with limited resource settings. Even in the most recent study [25], which a comprehensive analysis of Deep neural models with precisions over 97% has been assessed on G-banded images, the chromosome dataset is 7 times larger than ours, with 4 times higher number of G-band levels (300 vs 72). To provide more details on the proposed metrics effectiveness, we rather focused on comparing proposed metrics in variety of conditions like feature types, data pruning effects and classifier linearity effect. Therefore, the baseline for comparison has not been selected from other methods, and merely simplified classifiers like Alex-Net and SVM are used.

New metrics have been proposed in this paper to detect the trained classification model limitations and uncertainties in a label-dependent manner. There are varieties of factors playing roles in the mistakes and therefore uncertainties classifiers cause in testing phase, e.g., closeness to decision region, inference on small sample size in region of interest, noisiness of data, and also model imprecision in attributing close scores to more than one label. To minimize the chances of model bias effects on test data while avoiding the model to overfit the train data, simpler models can be trained and then passed through a label-dependent thresholding process in which thresholds are found in such a way to have the highest validation data Precision per label. This approach, reinforced by our proposed metrics, aims to detect where the classifier is uncertain and tackle models that suffer underfitting issue in specific contexts and labels. The evaluations have been implemented on the meta-phase image data under raw, scale invariant and engineered features. Moreover, the proposed effectiveness approach has been assessed in different linear and nonlinear classifier settings, with and without dimensionality reduction and data pruning. The purpose of pruning data is to avoid misclassification outcomes out of highly curved data in the database due to their high variability. Before passing to chromosome identification classifier, data is pruned by extracting straight or semi-straight images from overlapped and corrupted kinds, to then be fed by a 24-class chromosome classifier. This pruning approach improves the Precision of the estimated labels even more, without pruning a large amount of the data (i.e. 10-20%).

The purpose of this paper is to address the reliability and certainty of G-banded chromosomes classification to provide high Precision for them. Therefore, the main contributions made in this paper are as follows:
- Improving the Precision of low-quality semi-automated Karyotypes using automated threshold metrics.
- Using invariant and engineered features suitable for identification
- Tuning and learning linear, neural, and supervised dimensionality reduction models to improve Precision.
- Training pruner classifier to remove uncertain test inputs.

The structure of the paper is as follows. The proposed thresholding and reliability metrics are described in Section 2.1. The next section deals with the input data types used in the utilized classifiers. Section 2.3 provides information and statistics about the meta-phase chromosome database. The classifiers implemented in this paper are described in Section 2.4 and the tuned hyper-parameters used in each has been mentioned. In the fourth subsection of Section 2, the proposed threshold metrics, pruning classifier, and chromosome sorting classifiers are applied to Karyotypes. The comparative results are shown in Section 3. In Section 4, discussions are provided about proposed metrics and classification results. At last, the conclusion closes the paper by prospective approaches.

## 2 Materials and Methods
### 2.1 Proposed reliability metrics and thresholding approach

The main objective is to improve the Precision per each label over unseen chromosomes by adding an extra label named "uncertain" to give the opportunity for classifiers not to classify in uncertain situations. As discussed in the introduction, such settings help the semi-automated softwares to delegate suspicious images to the operator and reduce necessities to review all the automated classifications rigorously.

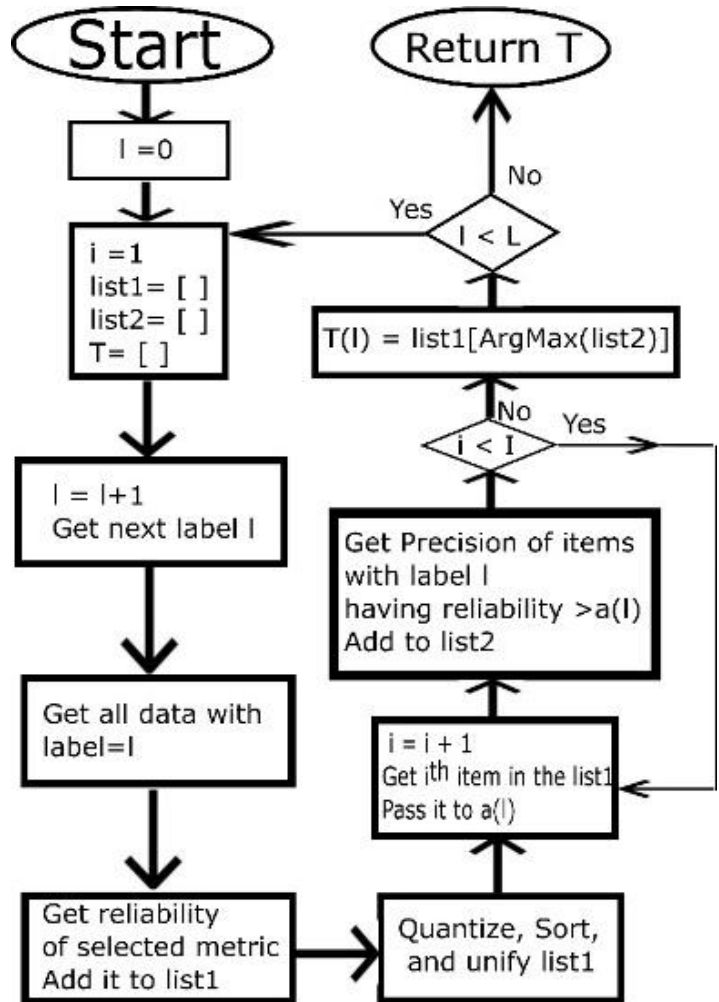

**Flowchart 1,** Flowchart of threshold approach per label using proposed metrics. Inputs are Trained Classifier, Train data the classifier is trained on, A preferred label index which demands a reliability threshold, A preferred classifier reliability metric from Table 1, and Eligible Recall threshold. Algorithm output is the learned thresholds for each label in the corresponding classifier.

Five different metrics have been proposed for measuring classifier certainty per label. Then, the values computed per instance for each correctly classified label are averaged to reach a finalized threshold for correct classification. The metric values over labels should be higher than the computed average to pass the certainty test and be regarded as a reliable decision. Table 1 discusses the proposed reliability metrics mainly to provide heuristics for assessing reliability and decision certainty of classifiers with output scores. Most of the classifiers return scores for each label which is Softmax output in discriminative neural networks, negative Hinge Loss in Support Vector Machines (SVM) and etc. However, our method is not applicable in fuzzy inference systems and similar rule-based methods. The columns mentioned in Table 1 respectively describe the metric name, metric formula, best sought performance so far by using such metric as reliability threshold, the classifier in which the best result is sought, number of all validations/labels having such metric used for thresholding, and the evidence of its practicality in assessing classifier certainty. The values resulted from these metrics help the semi-automatic system inform the operator about uncertain instances deserving extra attention. For evaluation and comparison of metrics with each other, the metrics performance on chromosome classification data mentioned in Section 2.3 on pruned data have been shown.
The performance is based on the Precision of label detection and also Precision improvement compared to case without reliability thresholding.
The main flowchart for training the labels' thresholds is proposed In Algorithm 1.

After learning the threshold values for each label, the classifier can be tested on a different validation or test data unseen by the model and threshold derivation phase. Algorithm 2 illustrates the instance verification approach used to assess a test instance:

**Table 1** Metric information and performance comparison on Section 2.3 dataset and pruned data 24-class classifiers in Section 2.4

| ID | Metric name | Formula | Best Prec. (Section 3) | Best Prec. classifier | Best Prec. Improve (Section 3) | Evidence used to derive metric |
|---|---|---|---|---|---|---|
| I | Classifier score per label | CNN: $\delta(W_l x_l^j)$ Where $\delta$ is the Sigmoid function, $W_l$ is parameters matrix of last layer (each row corresponded to 1 label), and $x_l^j$ is last layer input for jth batch.<br>1-vs-1 SVM: $S_i = \sum_j R(1+w^{l_i,j^T} x_i)$ Where R is Relu activation function, $w^{l_i,j}$ is learned hyperplane normal for i'th input x in a binary SVM classified w.r.t. j'th label, $l_i$ is class label index for i'th input. | 0.99576 | SVM | 0.00140 | The higher the score, the farther the instance is from the decision region of a well-trained classifier. |
| II | Score divergence from top 4 classifier score average | $\|S_{est} - S_{top4}\|$<br>$S_{est}$: score of the estimated label<br>$S_{top4}$: Average over top 4 maximum scores except for the estimated label score | 0.81033 | ANN | 0.06033 | The higher the score divergences, the higher the division of the estimated score from those of other labels'. |
| III | Two highest scores difference | $\|S_{max} - S_2\|$<br>$S_{est}$: estimated label's score<br>$S_2$: maximum score regardless of the estimated label's score | 0.97753 | KNN | 0.93587 | The likelihood of confusing one label with another most likely case by a classifier. |
| IV | Score variance over all labels | $Var(S)$<br>S: vector of scores for all labels | 0.97753 | SVM | 0.76483 | The more variated the scores are, the more discriminating the classifier behaves. |
| V | Estimation score & min. score difference | $\|S_{est} - Min(S)\|$<br>$S_{est}$: estimated label's score<br>S: vector of scores for all labels | - | - | - | - |

**Input:** Pretrained Classifier/ Unseen validation or test data/ preferred classifier reliability metric (Table 1)/ derived reliability thresholds
**Output:**
Boolean value stating the reliability of decided label
**Main Algorithm:**
Classify the test instance with the pretrained classifier
Get the score vector of the classified instance
Compute the metric output over the score vector for the estimated label's score
If Computed_metric > Reliability_threshold(estimated_label):
    Return True
Otherwise:
    Return False

**Algorithm 2** Pseudo-code of reliability assessment of each decision made by the classifier

The objective of reliability testing approach is to classify chromosomes. Therefore, the comparisons and evaluation results have been provided in Section 3, using SVM and ANN as baselines for linear and nonlinear classifiers. The method improves Precision substantially on unseen test data as it assigns "uncertain" label to most of the uncertain instances.

## 2.2 Input data and features

In this section, the types of data fed into the classifiers have been described. The methods are divided into three main types, i.e. resized images fed into CNN, features derived by Scale Invariant Feature Transform (SIFT) for SVM, and features engineered corresponding to bands of chromosome, its relations, and intensity profiles also for SVM classifier. For SVM, they either have to get reduced scale and rotation invariant features provided by SIFT, or need to have selected features computed.

The chromosome database introduced in Section 2.3, has to pass through varieties of preprocessing methods to get ready for the training phase for each of three aforementioned input types. First, the metaphase image is sent to an automatic chromosome segmentation phase described in Algorithm 3. Then, the middle line's tangent per each segmented chromosome is extracted using Algorithm 2 to rotate the chromosome to a vertical posture. Finally, the images are saved for each of the following input schemes.

Before extracting the features, all chromosomes are rotated to vertical instance having p-bands upward. Because approximately all the chromosome indices have q-bands lengthier than p-band, centromere distance to top and bottom of chromosome are used to rotate p-band upward. To compute the centromere location, first 10-peakiest locations are sought, then three highest boundary curvature points are selected from left and right side of chromosome boundary (as dissected by the blue line in Fig 2) and the point with highest peak is selected out of them. The method for finding curvature is discussed in Section 3.3.1 in part named "Curvature Profile". The selected points, one from right-side-boundary and the other from left, are usually aligned to each other. Otherwise, the algorithm is set to return no boundaries.

The boundary finding process specialized for dataset in Section 2.3 is described in Algorithm 3. The algorithm is derived empirically and works for all train data in Section 2.3.

**Input:** Grayscale image
**Output:** Boundary points
**Algorithm:**
Normalize image.
Filter image with Gaussian method and variance 1.
Threshold image intensities under 220/256 to be 1, otherwise 0.
Impose closing algorithm on the threshold by a disk of size 2.
Compute gradients along x and y using Sobel Edge Detector (SED) with a 3-sided Sobel Kernel.
Compute the image gradient intensity by adding squares of x and y gradients together
Threshold image intensities under 150/256 to be 1, otherwise 0.

**Algorithm 3,** Process of finding chromosome boundary

### 2.2.1 Input for ANN

The first input type is mainly passed through a pipeline of grayscaling, normalization and Histogram Equalization (HE) process for a more homogenous lighting condition in each image region. For a description about HE, the reader can refer to [9]. Due to the fact that images have varieties of shapes and sizes, they all have been resized to a specific size of 200x100. The resize rendering method was set to nearest neighbor.

### 2.2.2 The preferences for SIFT

SIFT is a method to capture features robust to rotation and scaling [10]. First, it uses group of Laplacian of Gaussian (LoG) filters with varieties of window width for image filtering to find image gradient information in a multi-scale way. Then each LoG result seeks for keypoints where their gradient intensity in all scales are the highest of all. To filter more prominent intensity values, a threshold is used to prune low gradients and afterwards the histogram values of the intensities are added to the feature descriptor vector as finalized features. This process is repeated for each sought keypoint and for different orientations to construct the final feature vector. For the detailed information, the reader can refer to [10].

The main input parameters for SIFT are number of number of keypoints and number of histogram orientations. To suit for the dataset mentioned in Section 2.3, these parameters are fine-tuned from the sets of {5, 25, 50} and {32, 64, 128, 256} respectively. The optimal values for number of keypoints and orientations on train data generalization error are 50 and 128 respectively.

### 2.2.3 Feature engineering methods suited for G-banding

Two types of features are extracted to better represent characteristics of the chromosomes, i.e., morphological and structural. Because the chromosome sizes vary, the provided features are scale independent and have a relational value. Morphological features are usually lengthy and represent the shape and intensity statistics in this context, while structural features capture the banding length relations and other centromere information to focus on chromosome characteristics. The overall number of features are 2166 for evaluated dataset in Section 2.3.

### 2.2.3.1 Morphological features

Many trials and errors have been approached to finally select and put the appropriate features together. The following feature extraction schemes have been ideated from the literature [1, 2, 3, 11, 12, 19, 21, 22]. The features in this group are representing shape and profile information of chromosome evident in the morphology.

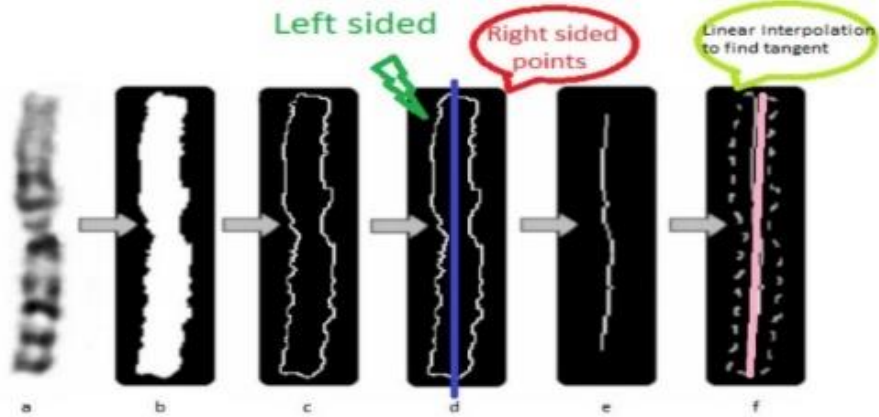

**Fig 2** the process for computing tangent value.

**Tangent of Middle Line**
- To find the middle line of a vertical chromosome, the points from the left side of chromosome boundary are averaged by their counterparts on the right boundary side.
- To find the tangent, a Least squares (LS) estimation is used to find **x_1** in A**x**=b where x=[$x_1$,$x_2$] is [$m_{tangent}$, $y_{intercept}$], $A_{n\times 2}$ has two columns being middle line's z coordinates and 1, and finally b is a vector of middle line's y coordinate. Using LS , **x** is $(A^T A)^{-1} A^T b$.
- The process of boundary extraction, middle line finding, and tangent computation is summarized in Fig 2.

**Intensity profile**
- To compute intensity profile, per each row in vertical chromosome, the average of all intensities from the point to the left side of boundary to the right side is computed and appended into a vector.
- As Fig 3 shows, the yellow star represents left side boundary point while the blue star is pointing to the right side boundary in the corresponding row. All image intensity values per each red line are averaged together and the finalized vector of profile is recorded.
- The number of items determines the distance of contiguous red lines and are set to 334 for all images to account for different scales. The number matches the height of scaled chromosome images used in the neural network phase during evaluation.

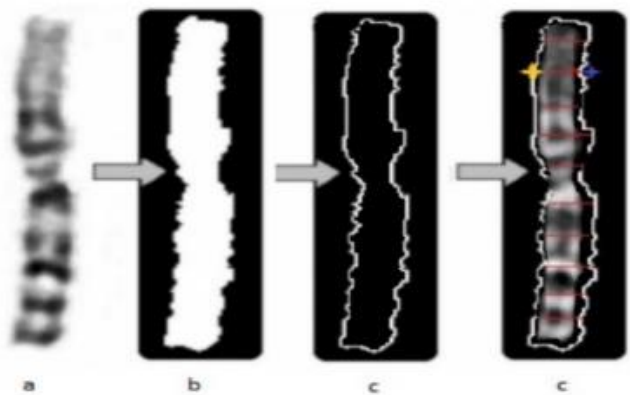

**Fig 3** the process for computing intensity profile.

**Width profile**
- It is a vector out of lengths of each red line in Fig 3.
- The vector length is 334.

**Curvature profile**
- To compute curvature per different rows of vertical chromosome, second order polynomial curve fitting can be used. Similar to the tangent finder, LS method is needed for approximation. The number of curvatures is set to 30

**Shape profile**
- This method, does a weighted average of rowwise point distances from middle line weighted by the intensity values. The middle line is shown in Fig 2 part f in pink. The number of elements in the resulted vector is the same as other profiles.
- The formula for computing each value is shown below in Formula 1.

$$P_j^{shape} = \frac{\sum_{i=1}^{n} g_{i,j} \times d_{i,j}^2}{\sum_{i=1}^{n} g_{i,j}} \tag{1}$$

Where $g_{i,j}$ is the i'th chromosome point's intensity per row j, and $d_{i,j}^2$ is its distance from the middle line.

**Weighted intensity profile**
- This method, does a weighted average of rowwise intensity values weighted by the point distances from middle line. The middle line is shown in Fig 2 part f in pink. The number of elements in the resulted vector is the same as other profiles.
- The formula for computing each value is shown below in Formula 2.

$$P_j^{w\_intensity} = \frac{\sum_{i=1}^{n} g_{i,j} \times d_{i,j}^2}{\sum_{i=1}^{n} d_{i,j}^2} \tag{2}$$

Where $g_{i,j}$ is the i'th chromosome point's intensity per row j, and $d_{i,j}^2$ is its distance from the middle line.

**High-peak features**
First distances of all points at the right and left sides of boundary to the middle straight line are computed from the vertical line shown in blue in Fig 2 part c, leaving 668 values. Then, top 334 highly peaked points corresponded to the distances are extracted from the boundary points. To find peaks, the points intensities increasing w.r.t. the previous point and next one are selected.

**Width variance**
The variance of the width-profile vector

**Height variance**
The variance of the height-profile vector sought by extracting width profile on transpose of chromosome image.

### 3.3.2. Structural features
21 structural features have been extracted, each manifesting specific discriminative aspect of chromosomes or the factors lie in identification of defects. These features' types are all scalers.

**Sum of proportions**
Bands' relative distances have a substantial role for operators to identify chromosome label in G-banded images. To account for multiple scales, the following formula is used for this feature computation:

F_s1 = ($d_{uq}$ / $d_{dq}$) / ($d_{up}$ / $d_{dp}$) (3)

Where:
$d_{uq}$ / $d_{dq}$ / $d_{up}$ / $d_{dp}$ is distance of the highest peak in Q / Q / P / P band to the chromosome top/ bottom/ top/ bottom,
P band is the band above the centromere position. Q band is the band below the centromere position.

**Proportion of Q to P length**
Proportion of the chromosome height below centromere to the height above the centromere.

**Relative chromosome length**
Sum of intensity values on the middle line to square root of image intensity sum

**Standard deviation below centromere**
In between centromere and the closest peak to centromere, the variation of chromosome image intensities can provide informative clues regarding the genetic abnormalities.
The process for finding the feature is:

**Density near centromere**
Average of all intensities from closest peak above centromere to closest peak below centromere, then dividing it by whole image's average.

**Chromosome length1**

The height of vertical chromosome image.

**Chromosome length2**

The normal chromosome length may sometimes be wrong especially for bent chromosomes. To tackle such issue, half of chromosome boundary length can be approximately used as chromosome length, provided that the value will be not affected by chromosome image scaling. To account for variety of image scaling, the result gets divided by chromosome width. Therefore the feature value is:

Number of thinned boundary points divided by chromosome thickness.

Where chromosomes thickness is average of width profile.

**Highest peak relative distance**

Distance of highest peak along column from bottom to its distance from top for a vertical chromosome image.

**Centromere curvature**

Curvature of 10 points near centromere row in the boundary, for the right side and left side differently. Curvature is found by LS method the way it is mentioned above for computing "Curvature profile" vector.

**Centromere width**

Distance of centromere column in the left side to centromere column in the right.

**Chromosome area**

Sum of the width profile vector

**Chromosome thickness**

Average of the width profile vector

**2.2.4 Use of supervised Dimensionality Reduction (DR) to reinforce linear classification**

The current studies have shown the effectiveness of dimensionality reduction using Principal Component Analysis (PCA) and wrappers. For instance, Lerner et.al. improved chromosome classification of ANN from 77.8% to 85.3% by using preprocessed PCA features [16]. However wrappers need Genetic Algorithm and can be computationally intensive and PCA has not provided satisfactory results on our dataset probably because of lack of supervision by discriminative power.

To utilize the SVM discriminative capability in a way to prevent it from undergoing the Curse of Dimensionality, the SIFT dimensions have been reduced to a lower subspace using Fukunaga transform [30]. The transform is based on Fisher LDA. However, it projects data to least within-class dimensional variance subspace and then projects the derived subspace to another lower subspace with high between-class variance per dimension [9]. The purpose is to reduce the division-by-zero effects of singular within-class covariance matrix on overall results of LDA. Formula 4 describes the process.

$$X_{final} = W_{between} X_{within} = W_{between} W_{within} X_{in} \qquad (4)$$

Where $X_{in}$, $X_{within}$ and $X_{final}$ are input data, transformed data in first stage, and final discriminative subspace; $W_{within}$ is rowwise catenation of column eigenvectors with least eigenvalues in $\overline{X}_{in}^T \overline{X}_{in}$, and $W_{between}$ is rowwise catenation of column eigenvectors of highest eigenvalues in $\sum_{i,j|l_i \neq l_j} (X_{within}^i - X_{within}^j)^T (X_{within}^i - X_{within}^j)$. The results on chromosomes dataset show the effectiveness improvement of dimensionality reduction compared to cases without it.

**2.3 Dataset used for classification**

The purpose of the proposed models and metrics is to improve the Precision of G-banded Karyotyping. Therefore, a medium to low resolution dataset have been used to assess the performance of the classification. To focus on model improvements for low-budget facilities, one of the G-banding datasets from an outdated low resolution Karyotyping system in Shiraz Jahad Daneshgahi Pathological Laboratory has been selected refraining to use the new existing systems there. The system name was VideoTest, updated in 2009, with software "Karyo 3.1", a 1998 version. The data has been collected from a 60DPI camera having raw metaphase image dimensions of 1044x971 pixels formatted as TIFF. The camera model used for taking picture was Nikon Eclipse 50i with 100X zoom power. To reduce pictures resolution fitting lower cost cameras, the images were decimated to 703x991. 2000 images have been extracted, passed to automatic chromosome segmentation, resulting 12513 chromosomes with the maximum height of 334 and maximum width of 306 pixels with the average height and width of 72.04, 77.42. The noises and the low resolution of data, reduced the true resolution by factor over 4, which leads to highly blurred images. After cropping, the preprocessing method mentioned in Section 2.2 has been imposed on the chromosomes solely. Afterwards, they are labeled to the indices from 1 to 24 which indices 1 to 22 manifest chromosomes 1 to 22, index 23 refers to Y and 24th index refers to X. An expert has verified the protocol and procedure of dataset recording. The chromosomes belong to male subjects for 64% and 36% to female ones. On average, the number of images per each subject is 6.7 and totally 997 raw images exist in the dataset each one containing at least 5 and at most 24 chromosome segments.

For the phase of data pruning, the pruner classifier had only four classes, named nearly straight, curved, overlapped, and noisy data. The data has been segmented to three groups of train, validation, and test with the proportion of 60%, 20%, 20%. To make the results realistic, the subjects used in each segment are different.

For the chromosome index classification model, data has been dissected to only train and validation with the proportion of 80% and 20%. 20% of images are highly curved or overlapped. What is more, they usually demand an automatic pruning phase to prevent accuracy fall. Therefore, the automated pruner classifies the semi-straight chromosomes. However, due to the 20% reduction of train and validation data after automated pruning stage, usage of test data are ignored to raise number of training and validation data, ensuring results that are more reliable. To make the results realistic, the subjects used in each train and validation phase, are set different from each other.

For 24 class classification scenario, the data is imbalanced (as Table 2 shows). Therefore, it demands reweighting approach while using SVM. The lines in blue are associated to chromosomes that the mentioned classifiers (in the next sessions) provide promising results in a low-quality dataset, i.e. over 80%.

Fig 4 shows the distribution of 24 classes in 24 different progressive colors before and after semi-straight data pruning and filtering. The visualization, provided by t-SNE approach (Matten and Hinton, 2008), shows that pruning provided by the pruner classifier (over data in Table 3) made same label data closer to each other. This effect helps improve classification performance. After pruning, the 2D visualization for labels especially those with color red and blue, tended to incline closer to their own kinds and get more compact.

For a sample from the evaluation dataset, the reader can refer to Fig 1. Fig 5 shows 4 samples from different labels gathered for training the semi-straight chromosome pruner. Using the pruner classifier is mainly for removing labels 1, 2 and 4 from data which were only 12.2% of whole in dataset and generally in the upcoming samples.

**Table 2,** Chromosomes distribution in 24 class classification

| Label | %Train | %Valid. | Label | %Train | %Valid. |
|-------|--------|---------|-------|--------|---------|
| 1  | 5.8 | 5.8 | 13 | 3.9 | 3.9 |
| 2  | 6.3 | 6.2 | 14 | 4.3 | 4.3 |
| 3  | 5.9 | 5.8 | 15 | 3.0 | 3.0 |
| 4  | 5.5 | 5.5 | 16 | 2.1 | 2.2 |
| 5  | 5.4 | 5.7 | 17 | 4.0 | 4.1 |
| 6  | 5.8 | 5.7 | 18 | 1.9 | 1.9 |
| 7  | 6.1 | 6.0 | 19 | 1.4 | 1.4 |
| 8  | 6.3 | 6.4 | 20 | 2.3 | 2.5 |
| 9  | 7.0 | 6.9 | 21 | 0.2 | 0.3 |
| 10 | 6.3 | 6.1 | 22 | 0.5 | 0.6 |
| 11 | 5.2 | 5.1 | 23 | 0.2 | 0.3 |
| 12 | 5.7 | 5.6 | 24 | 4.8 | 4.9 |

**Table 3,** labels distribution in data provided for data pruner classification phase. Data contains labels "curved", garbage", "Semi-straight", and "overlap" respectively as 1, 2, 3 and 4.

| Label | Name | %Train | %Valid. | %Test |
|-------|------|--------|---------|-------|
| 1 | Semi-straight | 89.6 | 80.2 | 89.6 |
| 2 | Garbage | 4.1 | 5.8 | 4.1 |
| 3 | Curved | 5.8 | 13.0 | 5.8 |
| 4 | Overlap | 0.4 | 1.0 | 0.4 |
|   | %Overall | 20 | 60 | 20 |

## 2.4 Implemented classifiers

Two types of classifiers are used for three types of chromosome classification. The two types are one-versus-one SVM classifier and variation of CNN named Alex-Net [15]. The followed classification schemes are threefold; identification before curved and overlapped chromosomes removal, after removal, and the removal process itself. The probability scores passed to thresholding metrics (those of Algorithm 1) are selected from "predict_proba" function in Scikit-Learn Python toolbox [24]. For Alex-Net, scores are set as the output to the last layer before Softmax activation function. The SVM classifier module used was LIBSVM [5]. Other hyper-parameters, i.e. decision function type and SIFT orientation, are one versus one and 128 respectively.

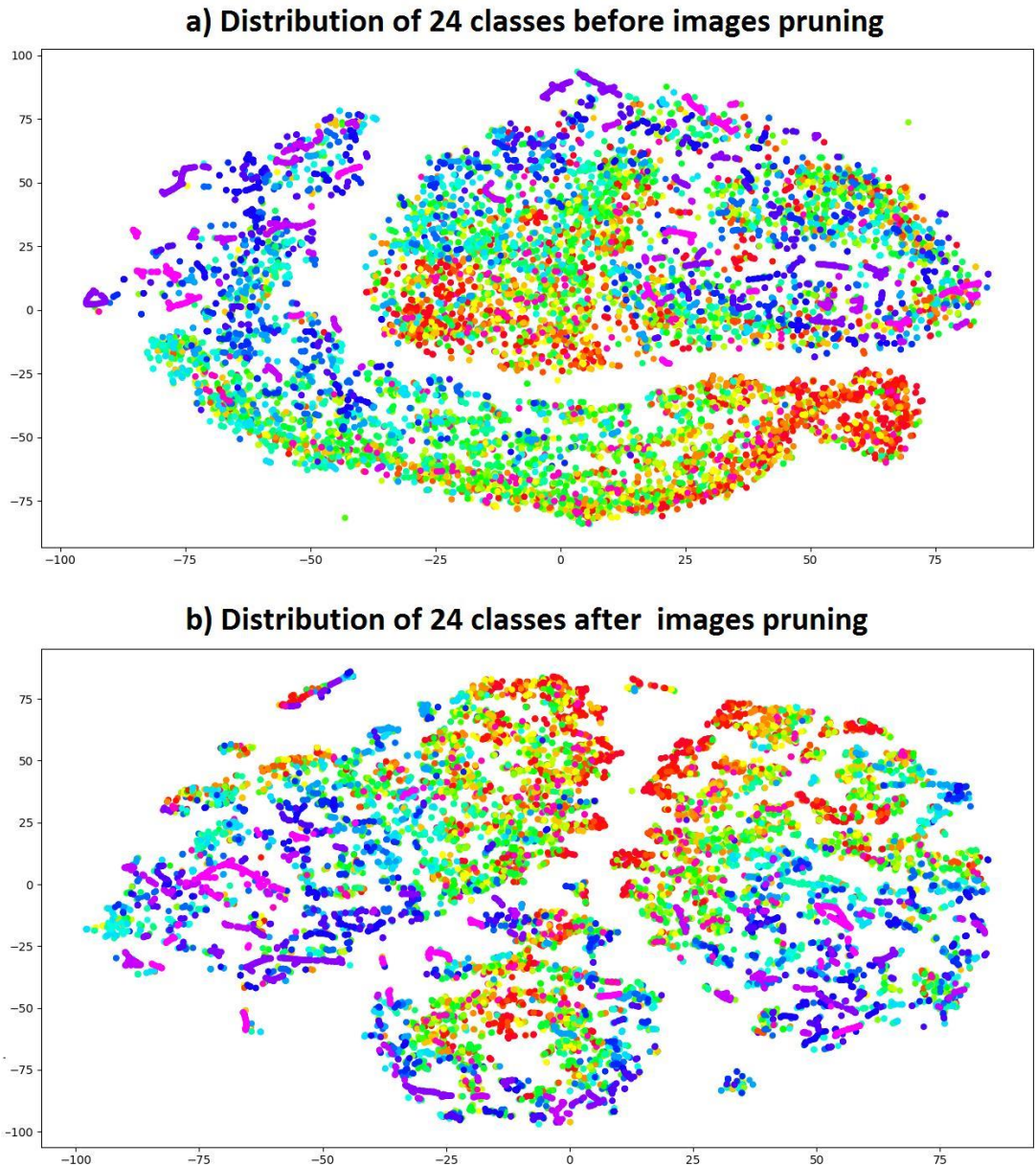

**Fig 4** t-SNE visualization per label after pruning curved and overlap chromosomes from the image.

Therefore pruning of them will not cause extra labor work for operators and instead provides a safe environment due to higehr likelihoods of curved items misclassifications. Then instances with label 1 are passed to the main 24-class classifier for chromosome identification. Table 3 shows the labels distribution used for train, validation and test data for the pruner classifier.

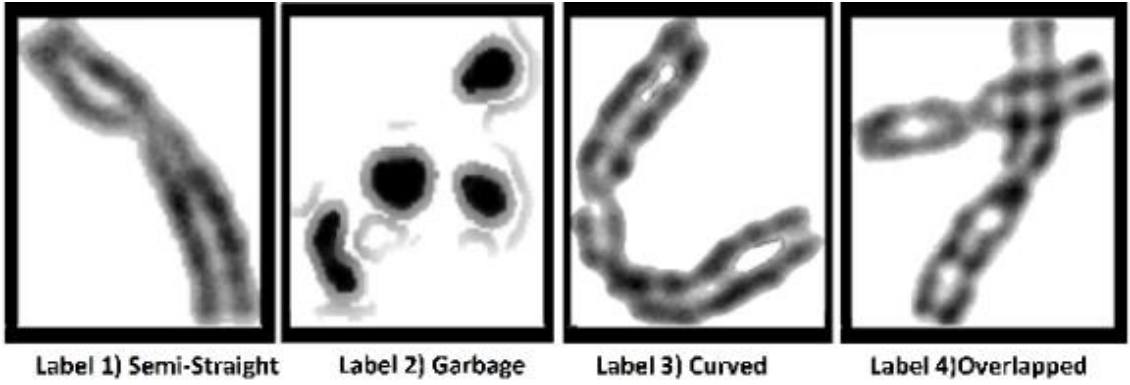

**Fig 5** The data pruner classifier dataset.

### 2.4.1 Neural network model for 24-class classification

In this stage, due to structure of images, the neural network used contains both convolutional and dense layers. It is a variation of Alex-Net with structure shown in Fig 6. The learning rate has been fine-tuned from the set {0.009, 0.015, 0.027}, maximum number of epochs are set to 60 during each run. Learning rate are adjusted every $l_a$ epochs reduced by factor $l_r$, which $l_a$ and $l_r$ are fine-tuned from sets {25, 15, 10} and {0.92, 0.87, 0.77} respectively. Batch size has been set to 20. Resulted hyperparameters $K_c$, $K_{d1}$, $K_{d2}$ are tuned to 32, 8, and 16 from {32, 36, 40}, {6, 7, 8}, and {12, 14, 16} respectively.

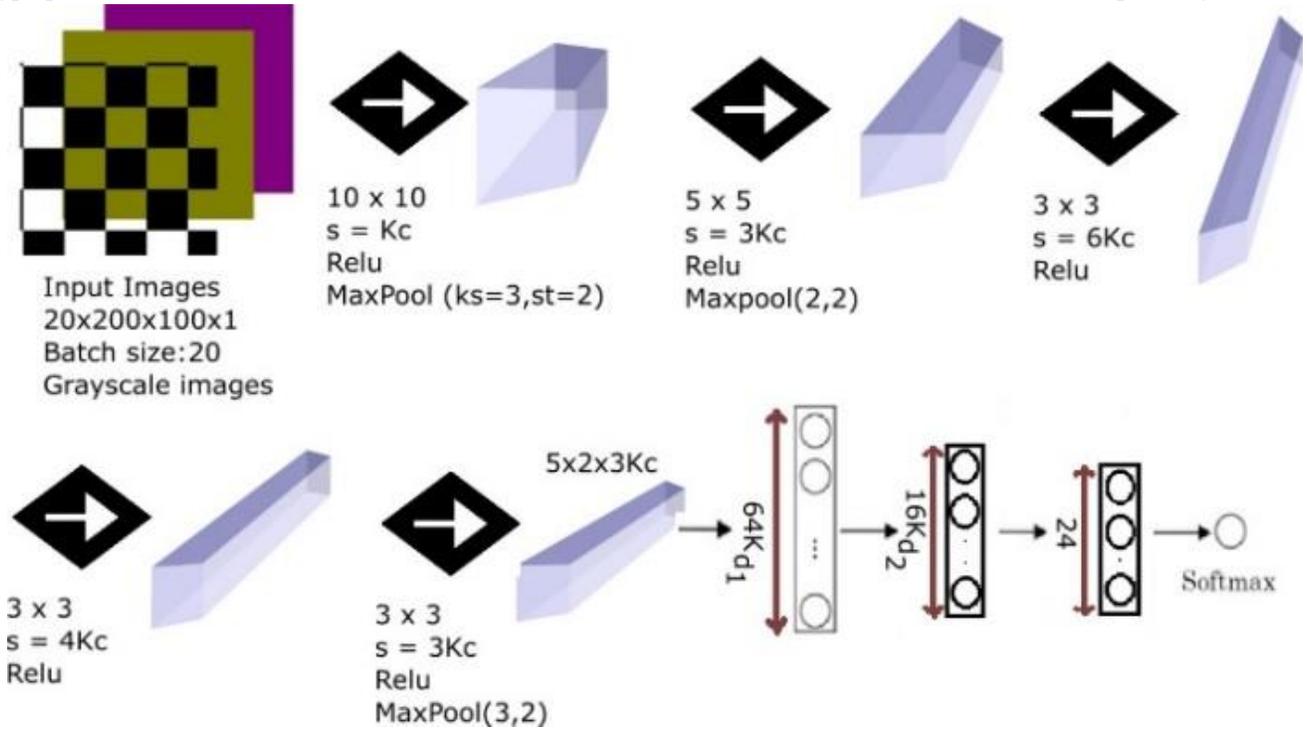

**Fig 6** implemented Alex-Net architecture

## 3 Results

The dataset mentioned in Section 2.3 has been used and the classification performance over the aforementioned cases has been analyzed. The performance metrics used in the Tables are Precision and Recall derived by the following formulas:

$$P(i) = \frac{C_{ii}}{C_{i,:}^T I(i)} \tag{5}$$

$$R(i) = \frac{C_{ii}}{I(i) C_{:,i}^T} \tag{6}$$

Where C is Contingency Matrix with rows associated to true labels and columns representing estimated labels. P(i) is Precision of label i, while R(i) gets the Recall the label i has. I(i) is vector with all values of one.

Cross validation has not been selected for a comparison metric in this study; because first chromosomes are multiple per each subject and random selection of chromosomes from the dataset reveals information about test data, therefore making the evaluation unreliable. Secondly, the subjects do not have similar number of segmented chromosomes failing to provide a valid subject-level cross-validation.

Table 4 provides the classification results of the data pruning classifier discussed and outlined in Section 2.3 and Fig 5. It shows the results of pruning improved compared to state without threshold metrics. Though this raise of Precision happens at the cost of Recall reduction, but such improvement could not even take place by merely observing the probabilities itself (i.e. index I) and therefore development of such relational metrics (i.e. in Table 1) has been helpful. Higher Precision values of data could also be possible by reducing the Recall cut-off in Algorithm 1. By creating such highly accurate pruner, it is hoped that overall reliability and Precision and therefore labor works of images reassessment are reduced. For the pruning classifier, only SIFT features are used with SVM classifier. Other parameter specifications are as mentioned in 2.4.2. To compare the results, metric indexed V, estimation score minus least score provided the best Precision among other metrics.

**Table 4** Precisions of pruning semi-straight chromosome images.

| Index of reliability thresholding metric | Test Precision using DR | Test Recall using DR | Test Precision without DR | Test Recall without DR |
|---|---|---|---|---|
| **I** | **0.94101** | 0.89001 | 0.93008 | 0.92681 |
| **II** | **0.93926** | 0.86592 | 0.93000 | 0.91716 |
| **III** | **0.93926** | 0.86130 | 0.93005 | 0.94933 |
| **IV** | **0.93926** | 0.86240 | 0.93008 | 0.94933 |
| **V** | **0.94358** | 0.86240 | 0.93008 | 0.95174 |
| **W/O reliab. thresh.** | 0.93926 | 0.91715 | 0.93008 | 0.95156 |

**Table 5** Precision and Recall results out of chromosomes identification process when the considered Alex-Net structure is used as its classifier.

| L | Pruned data Prec. by reliability metric | Recall for pruned data using reliability metric | Precision for pruned data without reliability metric | Reliability metric, pruned data | Prec. before prune | Recall for unpruned data | unpruned data Prec. without reliability metric | Reliab. metric, unpruned data | Model parameter specfication |
|---|---|---|---|---|---|---|---|---|---|
| 1 | ***0.92992*** | *0.71240* | *0.91724* | *III* | 0.70000 | 0.57534 | 0.70000 | - | $L_r$=0.009, $l_a$=50, $l_r$=0.75 |
| 2 | ***0.92827*** | *0.36330* | *0.80000* | *III* | **0.66170** | 0.02942 | 0.59259 | II | $L_r$=0.009, $l_a$=50, $l_r$=0.75 |
| 4 | ***0.89015*** | *0.79174* | *0.80662* | *III* | 0.74834 | 0.80142 | 0.74834 | - | $L_r$=0.015, $l_a$=50, $l_r$=0.75 |
| 5 | ***0.82341*** | *0.76096* | *0.62832* | *III* | **0.73311** | 0.60342 | 0.60396 | II | $L_r$=0.009, $l_a$=50, $l_r$=0.75 |
| 12 | ***0.81033*** | *0.79960* | *0.75000* | *II* | **0.76325** | 0.45100 | 0.75248 | I | $L_r$=0.015, $l_a$=50, $l_r$=0.75 |
| 18 | ***0.82397*** | *0.71028* | *0.75701* | *III* | **0.71840** | 0.59860 | 0.67333 | IV | $L_r$=0.015, $l_a$=50, $l_r$=0.75 |
| 19 | ***0.93026*** | *0.68505* | *0.74444* | *III* | **0.68801** | 0.45390 | 0.64646 | II | $L_r$=0.015, $l_a$=50, $l_r$=0.75 |
| 21 | ***0.90177*** | *0.45540* | *0.77311* | *III* | **0.73315** | 0.46280 | 0.70192 | III | $L_r$=0.015, $l_a$=50, $l_r$=0.75 |

Table 5 shows the ANN results whose model has been mentioned in Section 2.4.1. ANN can detect chromosomes 1, 2, 4, 5, 12, 18, 19 and 21 with high Precision. The chromosome indices failed to provide satisfactory result and improvement (i.e. higher than 80%) has not been mentioned in the table. The results regarding the pruned data, i.e. after removing curved overlapped images, has been shown in italic while results before pruning are shown in simple form. The comparisons of column 2 with 6 from this table shows that pruning helped improve the Precision substantially without loss of Recall except for chromosome labels 4 and 21 in a very small amount (i.e. only 1 percent). Moreover, Precision values before and after imposing reliability metric thresholds have been compared in two cases, first for pruned data in between column 2 and 4, and secondly before pruning between columns 6 and 8. The reliability metric thresholds used in this process have been derived using only chromosomes train data. They are evaluated on validation data described in Section 2.3. The bold values represent improvements caused by the trained reliability thresholds. Before pruning and filtering semi-straight chromosomes, the reliability thresholds seem ineffective for chromosomes 1 and 4 and therefore no Precision improvements are seen about them.

However pruning not only caused these two metrics to improve the Precision significantly by themselves, but also has given the chance for reliability thresholding metrics to improve Precision even more. Each threshold index responsible for the best Precision result is mentioned in the fifth and ninth column of Table 5 in Latin indices, being equivalent to the indices in Table 1 first column.

The results have also been improved using engineered features on SVM and KNN partly and are shown in the next paragraphs. To account for implausible results of ANN in other chromosomes and to improve the Precision for detection of other chromosome labels, a more deliberated feature extraction approach has been elaborated. The resulting features are passed to KNN. Table 6 shows the feature-engineered SVM chromosome classification results with and without including reliability metrics, semi-straight data pruning, and DR. The results in this table represent substantial improvements of Precisions for labels 1, 2 and 12 from 0.929, 0.928 and 0.810 to 0.977 and 0.977 and 0.995. Though the Recall values are so low, it shows further examinations may bring about a moderate Recall with higher Precision. Using proposed thresholding approach, none of the metrics found any low-Recall validation result with Precision higher than 95%. However, the proposed feature engineering approach could derive high Precisions over 97% for 5 chromosome indices. Rather than improved performance among 3 chromosomes, results got favorable for two other chromosomes labeled 3 and 7. Therefore, the SVM and KNN results once again verified the effectiveness of our proposed metrics for Precision improvement.

For the feature-engineered classification performances mentioned in Table 6, the results for classification before data pruning have not been provided. Because all the results were substantially fewer at least under 60% Precision. It verifies the fact that extracted features are not suitable descriptors for highly curved and overlapped images. All the features extracted for these results have undergone the supervised DR mentioned in Section 2.2.4, as the results without DR would not improve ANN. Between chromosome indices 1, 2, 3, 7 and 12, four out of five results of feature extraction using Section 2.2.3 features outperformed SIFT method. It suggests such extracted features are informative to uncover underlying discriminative information from the chromosomes.

Table 6  Classification results over feature-engineered data. Perfect Precision scores but at the cost of low Recall.

| Label | Precision for pruned data after imposing reliability metric | Precision for pruned data without imposing reliability metric | Reliability threshold metric for pruned data | Feature extraction type | Model parameter specification |
|---|---|---|---|---|---|
| 1 | **0.97753** | 0.31847 | IV | Morphological+Structural (Section 2.2.3) | KNN, n_neighbors=20 |
| 2 | **0.97753** | 0.35714 | III | SIFT (Section 2.2.2) | SVM, Keypoints: 50 |
| 3 | **0.97753** | 0.21276 | IV | Morphological+Structural (Section 2.2.3) | SVM, Linear kernel Keypoints count: 50 |
| 7 | **0.97753** | 0.04166 | III | Morphological+Structural (Section 2.2.3) | KNN n_neighbors=10 |
| 12 | **0.99576** | 0.99436 | I | Morphological+Structural (Section 2.2.3) | SVM, Linear kernel Keypoints count: 50 |

## 4  Discussion

### 4.1 Performance comparison of reliability metrics proposed in Table 1

To have a comparison between different reliability metrics mentioned in Table 1, each reliability metric has been evaluated by the pruned chromosome database in 24-class classification procedure in terms of :
- best validation Precision in all sessions
- best Precision improvement w.r.t. no reliability checking mode
- the number of labels the metric successfully improved in terms of higher Precision.

The performance results are shown in columns 4 and 6 of Table 1. Results from Table 1 show that metric III not only brought about the highest classification Precision of all the other metrics, but also provided the highest increase in Precision w.r.t. the classification without reliability metrics. The next favorable performance has been seen from metric IV with high final Precision. High performance improvement is compared to no-reliability-check-mode. The reason for outperformance of metric III and IV in comparison with other metrics are probably the following ones:

● Metric III, two highest score difference, is probably the best compliment for reducing common mistakes in some types of ANN and also KNN. ANNs may fail to best discriminate between the top classes densely concentrated in some regions. This effect makes it more likely to choose a wrong label over the other. Improving the model complexity also may not be the cure at such cases, as it may lead to over-representation issues at other regions despite even solving the underfitting problems at the region of interest. Therefore a reliability metric that compares and accounts only for first and second highly likely class labels, may outperform better than metrics considering other Sigmoid outputs. For the case of KNN, a similar occurrence may be the case. Because K in KNN is region-independent and not fine-tuned, it may bring about underfitting for some regions, and therefore may make mistakes in classification inevitable. In some datasets, a metric sensitive to the second most frequent class in region can reduce the possible classification mistakes more than metrics that regard many lower frequency classes having no substantial effects in classification over there.

● Metric IV, which is variance over all labels, has the second best performance. The reason may be the power of metric to measure the extent which the classifier discriminates between labels. Fig 7 suggests that the effectiveness of this metric is very low compared to metric III. Moreover the Recall provided by such metric is very low making it harder to be effective, especially when it failed to have favorable results in KNN and ANN as Tables 5 and 6 show. Though the metric can be a suitable tuner of classification performance based on train data itself without any needs to validation data. Ccontrary to other metrics, it does not depend on the estimated label score making the computation less time consuming.

● Among the other metrics, Score divergence from top (Metric II) had better Precision improvement than Classifier score (metric I). Hence, there are more discriminating information lying in scores' relational differences rather than in score intensities themselves.

● Metric V, maximum score and minimum score difference, did not lead to any promising performance and it seems that the minimum score does not have any suitable information to help regularize discrimination.

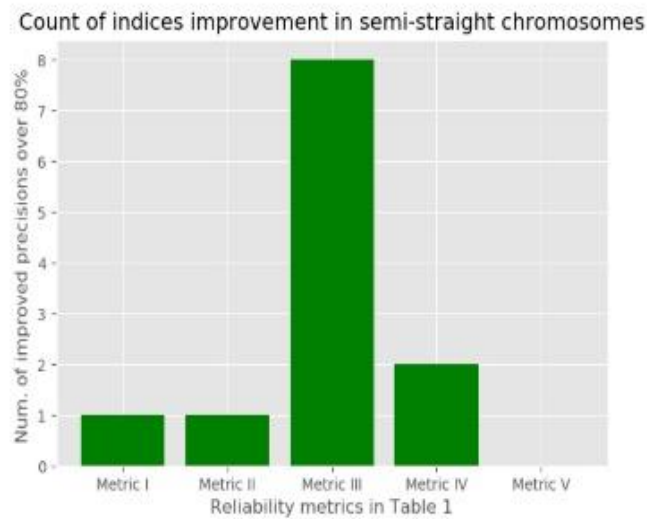

**Fig 7** Number of chromosome labels in the mentioned pruned-images 24-class classification that reliability thresholding process succeeded improvement whose Precisions are above 80%.

### 4.2 Discussion on Precision results

The utilized dataset Precisions concerning previous works have not been reported. The reason is the low performance results they provide for low-quality images dataset used in their settings and their classifier parameters. Even resizing the images and passing the dataset to the methods mentioned in the introduction did not provide satisfactory results (i.e. over 60% for any chromosomes). As the purpose of this paper is to assess reliability metrics and pruning on Precision improvement, evaluation results of such works in the literature have not been included here.

The results out of Table 5 and 6 show that all the implemented subprocesses contributed to classification improvement, i.e. providing more promising results. They brought about certainty to the operator working with semi-automated chromosome classification software on the status of identified and uncertain items. The feature extraction approach improved 3 classes performance and added two other classes to high Precision status over 95%. The supervised dimensionality reduction approach substantially improved four-class classification for dissecting semi-straight from other curved overlapped and garbage-like images. The reliability thresholding approach outperformed the results in all cases and types of classifications mentioned above. These approaches together had roles in the purpose of the paper for developing a precise reliable chromosome ordering method for low-quality images. As the classification has been approached on a simple CNN, other newly proposed CNNs can improve the results and provide better reliability in Precision values.

Due to computational resource limitations, all the experiments have been performed only for one set, and therefore no statistical analysis is available. However, due to multiple runs in different tuning settings and multiple parameter learning sessions for the tuning process, the performances are unlikely to be resulted by chance.

### Conclusion

In this paper, the objective was to improve Precision of low quality G-Banded chromosome data classification used in semi-automated chromosome ordering softwares for public low-budget pathophysiological laboratories. Therefore a set of new metrics for learning thresholds that detect reliability of classification are proposed for each label. Semi-straight data filtering, deliberated feature extraction, supervised dimensionality reduction, and Alex-Net based CNN are assessed on the data. The results show very high Precisions are possible without loss of Recall. They also represent that perfect Precisions over 99% can also be possible but with low Recall.

In the upcoming works, a new dataset with more balanced classes can be used to train a new model for detecting curved and overlapped chromosomes disregarded in this work. Furthermore, the curved and overlapped segment of data may be also fed to models to improve the overall classification performance of Karyotyping systems.

### Conflict of interest

The authors declare that there are no conflicts of interest.